\documentclass[10pt,twocolumn,letterpaper]{article}

\usepackage{wacv}
\usepackage{times}
\usepackage{epsfig}
\usepackage{graphicx}
\usepackage{amsmath}
\usepackage{amssymb}
\usepackage{booktabs}
\usepackage{epsfig}
\usepackage{subcaption}
\usepackage{multirow}
\usepackage{algorithm}
\usepackage{algorithmic}
\usepackage{paralist}
\usepackage{pifont}
\usepackage{hyperref}

\usepackage{enumitem}
\usepackage{newfloat}
\usepackage{listings}
\usepackage{color, colortbl}

\newcommand{\cmark}{\ding{51}}
\newcommand{\xmark}{\ding{55}}
\definecolor{RowColorCode}{rgb}{0.80,0.80,0.80}

\newcommand{\model}{SALAD}

\makeatletter
\newcommand\footnoteref[1]{\protected@xdef\@thefnmark{\ref{#1}}\@footnotemark}
\makeatother

\newcommand{\shorteq}{%
  \settowidth{\@tempdima}{-}% Width of hyphen
  \resizebox{\@tempdima}{\height}{=}%
}

\mathchardef\mhyphen="2D

\def\wacvPaperID{443} % Enter the WACV Paper ID here

\wacvapplicationstrack % Uncomment this line if you are submitting to the Applications Track.

\wacvfinalcopy % *** Uncomment this line for the final submission

\begin{document}

%%%%%%%%% TITLE
\title{\model~: Source-free Active Label-Agnostic Domain Adaptation for Classification, Segmentation and Detection}

\author{Divya Kothandaraman, Sumit Shekhar, Abhilasha Sancheti, Manoj Ghuhan, Tripti Shukla, Dinesh Manocha \\University of Maryland College Park, Adobe Research\\}
%\and

\maketitle
\thispagestyle{empty}

%%%%%%%%% ABSTRACT

\begin{abstract}
    We present a novel method, \model, for the challenging vision task of adapting a pre-trained ``source'' domain network to a ``target" domain, with a small budget for annotation in the ``target'' domain and a shift in the label space. Further, the task assumes that the source data is not available for adaptation, due to privacy concerns or otherwise. We postulate that such systems need to jointly optimize the dual task of (i) selecting fixed number of samples from the target domain for annotation and (ii) transfer of knowledge from the pre-trained network to the target domain. To do this, \model~consists of a novel Guided Attention Transfer Network (GATN) and an active learning function, $H_{AL}$. The GATN enables feature distillation from pre-trained network to the target network, complemented with the target samples mined by $H_{AL}$ using transfer-ability and uncertainty criteria. \model~has three key benefits: (i) it is task-agnostic, and can be applied across various visual tasks such as classification, segmentation and detection; (ii) it can handle shifts in output label space from the pre-trained source network to the target domain; (iii) it does not require access to source data for adaptation. We conduct extensive experiments across $3$ visual tasks, \textit{viz.} digits classification (MNIST, SVHN, VISDA), synthetic (GTA5) to real (CityScapes) image segmentation, and document layout detection (PubLayNet to DSSE). We show that our source-free approach, \model, results in an improvement of $0.5\%-31.3\%$ (across datasets and tasks) over prior adaptation methods that assume access to large amounts of annotated source data for adaptation. Code is available \href{https://github.com/divyakraman/SALAD_SourcefreeActiveLabelAgnosticDomainAdaptation}{here}
\end{abstract}
\section{Introduction}
\label{sec:introduction} 
Deep learning solutions for visual applications such as semantic segmentation \cite{vu2019advent,tsai2018learning}, image classification, and document layout analysis \cite{li2020cross,rusticus2019document} require a large amount of annotated data. Two popular trends to deal with the lack of sufficient annotated data are Domain Adaptation (DA) and Active Learning (AL). In Active Learning (AL)~\cite{ash2019deep,prabhu2020active,sener2017active,saha2011active}, the model mines and annotates samples within a fixed budget (\textit{e.g.} $5\%$ of the available corpus of unlabeled data \cite{fu2021transferable}) to maximize the models performance. Typical active learning strategies include modelling diversity and uncertainty for efficient sampling \cite{fu2021transferable,bouvier2020stochastic}. Domain adaptation \cite{vu2019advent} aims at transferring knowledge from a ``source'' domain to the ``target'' domain of interest.

An amalgamation of Active Learning and Domain Adaptation, Active Domain Adaptation (ADA) ~\cite{bouvier2020stochastic,su2020active,prabhu2020active} has explored the use of annotated ``source''-data from a related domain to transfer knowledge (via domain adaptation) to the new ``target''-domain dataset, within a fixed budget of annotating the ``target'' data. The biggest drawback of ADA is that it requires a large amount of annotated source data. This is prohibitive in scenarios like autonomous driving where the system only has access to a network \textit{pre-trained} on the source data. The source data itself is \textbf{unavailable} after the pre-training, due to privacy issues or storage constraints \cite{kundu2020universal,kurmi2021domain}. 
%However, in real world cases, the ``target'' domain, where the learnt model is being deployed, can have a different distribution than the ``source'' domain where it is pre-trained on. The ubiquitous problem of the distribution of the target domain being different from the source domain can be solved by domain adaptation. 
\subsection{Main Contributions:}

In this work, we focus on the challenging problem of Source-free Active Domain Adaptation, SF-ADA, where we have access to a pre-trained ``source'' network, but the source data is not available due to privacy concerns or otherwise. Further, an unlabeled target dataset, and a small budget for acquiring labels in the target domain is specified. Moreover, the target domain may also have a shift in the label space from the source domain. We propose \textbf{SALAD}, a generic novel framework for SF-ADA, which jointly optimizes sampling target data for annotation and source-free adaptation of the neural network to the target domain. 

\model~holistically addresses two key challenges of the problem setting via two complementary components: the Guided Attention Transfer Network (GATN) for knowledge transfer from the pre-trained network to the target domain and an active learning algorithm, $H_{AL}$, for mining samples from the target domain for annotation:
\begin{itemize}
    \item {\textit{Knowledge transfer:} GATN transfers relevant knowledge at the feature level from the pre-trained network to the target network (Figure~\ref{fig:overview}). GATN uses a transformation network to modulate the features of the pre-trained network, in alignment with the target domain, followed by guided attention for selective distillation to the target network. The target network guides the knowledge transfer via labeled samples selected through AL.} %We use an attention weighted L2 loss, to enable feature distillation from the pre-trained features to the target features. 
    \item {\textit{Mining samples:} The effectiveness of GATN depends on the samples mined from the target dataset. While it is important to choose samples that are similar to the distribution familiar to the pre-trained network, for effective knowledge transfer, we need to ensure that the chosen samples are informative to the network w.r.t. the target dataset. To do so, $H_{AL}$ fuses transferability from the pre-trained network, as well as uncertainty w.r.t. the target network.}
    %as well as uncertainty and diversity in the target domain. 
\end{itemize}

\model~has multiple benefits. (i) Knowledge transfer happens at the feature space (output of the network before the decoder). Thus, our architecture is label space-agnostic and can handle domains that contain different number and types of classes (ii) Our method is task-agnostic, and can be applied across various visual tasks. (iii) The source model is not required while testing and can be discarded after training. (iv) Our method does not attempt to emulate source data using generative approaches, which is common in task-specific source-free domain adaptation~\cite{kundu2020universal,liang2021source}. This makes our neural networks easy to train. 

We evaluate \model~across three tasks. On classification datasets (MNIST, SVHN, VISDA), we demonstrate that even without the source data, \model~performs similar to or better than the prior active domain adaptation methods~\cite{su2020active,prabhu2020active} that use large amounts of annotated source data. Next, we evaluate on MNIST under $2$ distinct cases of shift in output label space and show that \model~is able to achieve $99.4\%$ of the accuracy in case of no shift in the label space, thus establishing the effectiveness of our model in scenarios with label shift. Our experiments on the CityScapes dataset for semantic segmentation improves accuracy by $5.57\%$ over fine-tuning. We also highlight the benefits of \model~over other adaptation paradigms in Table~\ref{tab:cityscapes_sota}. Finally, we conduct experiments on adaptation for document layout detection from PubLayNet to DSSE, where there is a shift in label space. \model~ imparts a relative improvement of $31.3\%$ over fine-tuning of the target network on the small dataset.
\begin{table}
% \footnotesize
\centering
% \begin{center}
\resizebox{0.9\columnwidth}{!}{
\begin{tabular}{c c c c c}
\toprule
Problem & Src. Data& Src. Model & Lab. Tar. & Un. Tar. \\ 
\toprule
Semi Supervised DA (SSDA) \cite{wang2020alleviating} & \cmark & \cmark & \cmark & \cmark  \\
Unsupervised DA (UDA) \cite{tsai2018learning} & \cmark & \cmark & \xmark & \cmark \\
Source-Free DA (SFDA) \cite{kundu2020universal} & \xmark & \cmark & \xmark & \cmark \\
Active DA (ADA) \cite{su2020active} & \cmark & \cmark & \cmark & \cmark \\
\rowcolor{RowColorCode}
Source Free Active DA (SF-ADA) & \xmark & \cmark & \cmark & \cmark \\
\bottomrule
\end{tabular}
}
\caption{\small{\textbf{Problem Settings:} We highlight various domain adaptation settings. Src. Data, Src. Model, Lab. Tar., and Un. Tar. refer to abundant labeled source data, Source Model, Scarce Labeled Target Data and Unlabeled Abundant Target Data, respectively.}}
\label{tab:problem_settings}
% \end{center}
%\vspace{-25pt}
\end{table}

\section{Related Work}
There is significant prior work on active learning, domain adaptation, and active domain adaptation. However, to the best of our knowledge, there is not much prior work on an approach that can generalize to all tasks, is source-free, leverages pre-trained models for adaptation and exploits the flexibility to annotate a small subset of the ``target'' domain.

\noindent \textbf{Active Learning}
Active Learning (AL) aims to acquire a given small budget of labeled data while maximizing supervised training performance. Uncertainty-based methods select examples with the highest uncertainty under the current model~\cite{wang2014new,roth2006margin}, using entropy~\cite{wang2014new}, minimum classification margins~\cite{roth2006margin}, least confidence, etc. Diversity-based methods choose some points representative of the data, e.g. core-set selection~\cite{sener2017active,sinha2019variational}. Recent approaches combine these two paradigms ~\cite{ash2019deep,prabhu2020active,zhdanov2019diverse}. However, traditional active learning methods do not take advantage of readily available pre-trained models trained on large scale datasets.

\noindent \textbf{Domain Adaptation}
Domain adaptation aims to transfer the knowledge learned by a source domain model to an unlabeled target domain. Some of the existing works align feature spaces of the source and target domains by learning domain invariant feature representations by divergence-based measure minimization ~\cite{kang2019contrastive,long2015learning}, adversarial training~\cite{sankaranarayanan2018generate,shen2018wasserstein,tzeng2017adversarial}, source or target domain data reconstruction ~\cite{bousmalis2016domain,ghifary2016deep}, image-to-image translation~\cite{mao2018unpaired,hong2018conditional} or normalization statistics~\cite{maria2017autodial,li2018adaptive}. However, domain adaptation methods typically require access to annotated source data, and do not consider the possibility of annotating a small subset of the target domain.

\noindent \textbf{Source-free Domain Adaptation}
\cite{li2020model} introduced the paradigm of domain adaptation where source domain data is not available due to privacy issues and only a model pre-trained on the source domain data is available. Existing works employ a generative approach where the trained model is used to generate source samples using batch normalization ~\cite{ishii2021source} or energy-based methods~\cite{kurmi2021domain} for classification task \cite{xia2021adaptive,kim2020domain,agarwal2022unsupervised}. Others use a combination of distillation-based approach~\cite{liu2021source} or information maximization-based approach ~\cite{liang2021source}. However, these methods do not consider using active learning to boost the performance, and typically do not generalize across tasks. 

\noindent \textbf{Active Domain Adaptation}
Active domain adaptation aims to adapt a model trained on source domain data to target domain by annotating a fixed budget of target domain samples. \cite{rai2010domain} introduced the task of ADA with applications to sentiment classification for textual data. They proposed a method employing a sampling strategy based on model uncertainty and a learned domain separator.  More recently, \cite{su2020active} studied ADA in the context of CNN’s and proposed a method wherein samples are selected based on their uncertainty and targetedness, followed by adversarial domain adaptation. \cite{saito2019semi} proposed an algorithm that identifies uncertain and diverse instances for labeling followed by semi-supervised DA. \cite{zhou2021discriminative} proposed a three-stage active adversarial training of neural networks using invariant feature space learning, uncertainty and diversity-based criteria for sample selection and re-training. \cite{bouvier2020stochastic} addressed the problem of lack of guarantee of good transfer-ability of features in domain adaptation. However, all of the above works use source domain data, which is prohibitive in terms of data privacy. %In our approach, we introduce the task of source-free active domain adaptation wherein only access to a model pre-trained on source domain data is available.
%\subsection{Comparisons with Related Work}
%In Table \ref{tab:problem_settings}, we present various domain adaptation problem settings. In this paper, we propose an architecture for source-free active domain adaptation (SF-ADA). On one hand, SF-ADA can be seen as ADA \cite{prabhu2020active,fu2021transferable} without source data. It can also be seen as an amalgamation of the notions of SFDA \cite{li2020model,liang2021source} and AL \cite{bouvier2020stochastic}. SF-ADA requires \model~to address knowledge transfer from the pre-trained network and acquiring samples for annotation in a complementary manner. The design of GATN incorporates the transfer mechanism with spatial and channel attention \cite{woo2018cbam,fu2019dual}. We wish to reflect that our novelty is in the formulation of guided attention, rather than spatial and channel attention. Similarly, we wish to highlight that the AL heuristic $H_{AL}$ builds on the notion of entropy and gradient computation and is customized to the problem of SF-ADA.
\section{\model}
We propose a generic novel method, \model~, for the problem statement where we assume \begin{inparaenum}[(i)]
\item{a network, $N_{S}$, pre-trained on a source-domain, $S$ (the source data is not  available for adaptation) and}
\item{an unlabeled target domain $T$ from which we are allowed to annotate $B$ images.}
\end{inparaenum}
The goal is to mine $B$ images to facilitate adaptation from $N_{S}$ to a network, $N_{T}$ which learns robust task-specific features for the target domain.

\begin{figure*}
    \centering
    \includegraphics[width=0.8\textwidth]{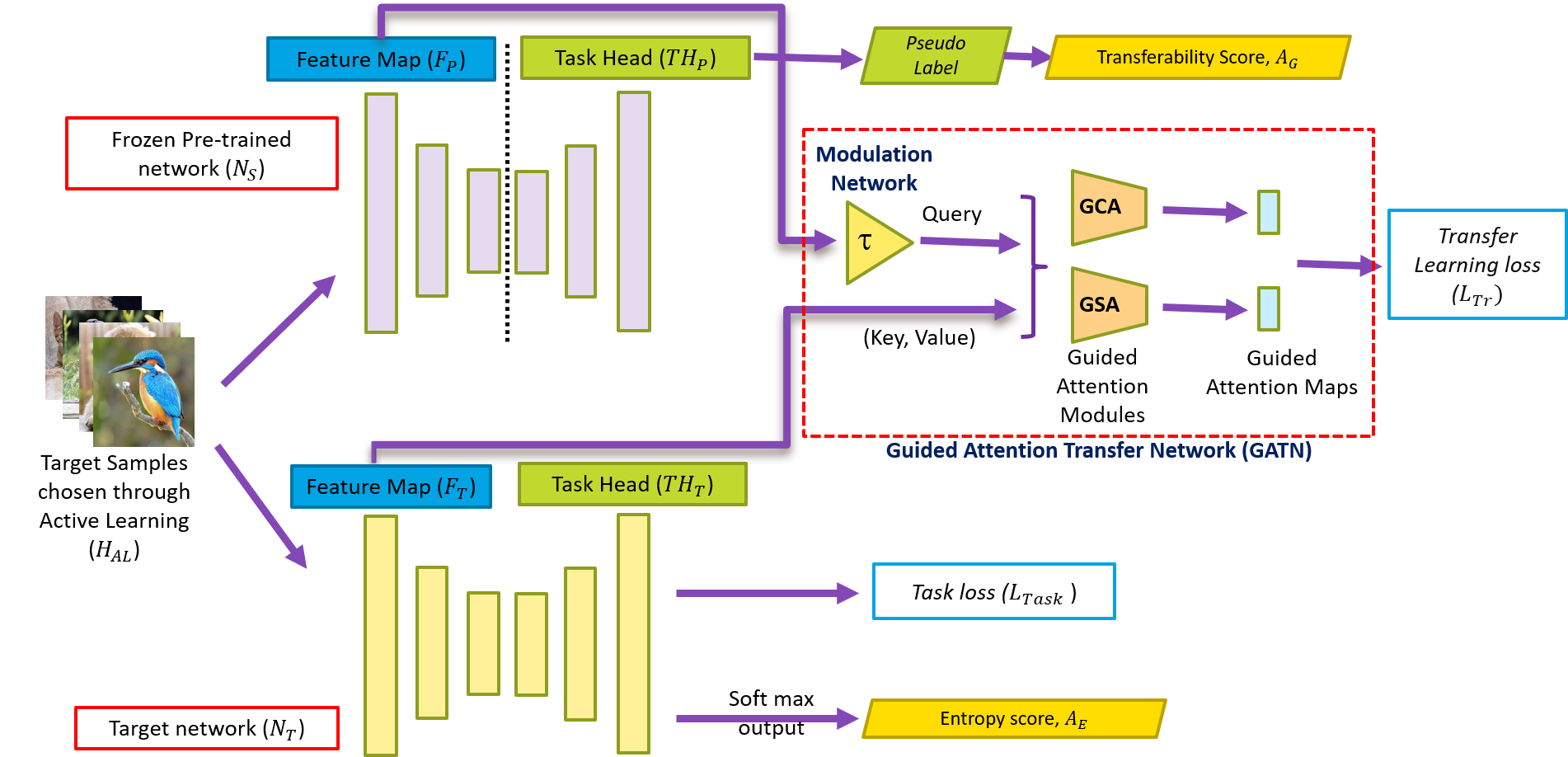}
    \caption{\small{We present a generic method, \model, for the task of adapting from a pre-trained source network to a target domain with a small budget for acquiring annotation, under a possible shift in label space. \model~consists of two complementary components: Guided Attention Transfer Network (GATN) and an active learning strategy $H_{AL}$.}}
    \label{fig:overview}
\end{figure*}

We postulate that methods designed for the above problem statement need to address two facets: \begin{inparaenum}[(i)] \item{Effective knowledge transfer from the pre-trained network to the target network using the small set of annotated samples and} \item{Intelligently choose samples to annotate, to facilitate knowledge transfer as well as span the spectrum of samples contained in the target dataset.}
\end{inparaenum} Consequently, our proposed solution, \model~ (Figure \ref{fig:overview}), consists of two complementary components: 
\begin{inparaenum}[(i)]
\item{Guided Attention Transfer Network (GATN) for knowledge transfer}
\item{an active learning strategy, $H_{AL}$, for acquiring samples.}
\end{inparaenum}
The goal of the AL strategy is to sample images that are important for the target network as well as facilitate knowledge transfer from the pre-trained network (without any negative transfer), while GATN strives to learn robust target domain feature maps by effective knowledge transfer from the pre-trained network to the target network. \model~ is a generic approach that can be applied across various visual tasks (classification, detection, segmentation). Moreover, it can handle shifts in label space from the source to the target domain where the source and target domains may not have the same set of classes. We now describe our method in detail. 

\subsection{Method Description}
Prior work on SFDA \cite{kurmi2021domain,liu2021source} has relied on the alignment between source and target data to guide the adaptation process. This requires the system to emulate source-data in a source-free setting which does not generalize across complex tasks \cite{liu2021source,kothandaraman2020ss}. Instead, we adopt a novel distillation-based transfer learning strategy at feature level. The selective distillation, combined with AL strategy $H_{AL}$, enables domain-specific learning for target network using chosen samples, where relevant domain-agnostic knowledge is transferred from the pre-trained source network via attentive distillation.

For initiating the distillation, the frozen pre-trained network, $N_{S}$ is split into a frozen feature encoder which generates feature maps $F_{P}$ and a frozen task head $TH_{P}$. Concurrently, the trainable feature encoder of the target network, $N_{T}$ is used to generate target feature map $F_{T}$ We depict the trainable task head of the target network as $TH_{T}$. $F_{P}$ and $F_{T}$ are passed through the Guided Attention Transfer Network (GATN), which constrains the target network via a transfer learning loss $L_{Tr}$. Our use of guided attention enables the target network to transfer the domain-agnostic features from the pretrained network, and discard domain-specific features from the pretrained network. The GATN, as well as the pre-trained network, can be discarded after training, i.e. they are not required in the evaluation phase.

The GATN consists of a modulation network, $\tau$, for transforming $F_{P}$ to modulate pre-trained network features in alignment with the target network. The transformed $F_{P}$ and $F_{T}$ are then fed to two guided attention modules to compute attention across spatial and channel dimensions. GATN enables the learning of transferable features for knowledge distillation. %The guided attention modules consist of Guided Spatial Attention (GSA) and Guided Channel Attention (GCA) \cite{woo2018cbam,fu2019dual} to compute attention across spatial and channel dimensions respectively. 
The overall adaptation framework (GATN and the target network) is jointly trained with a combination of the transfer learning loss $L_{Tr}$ and the task specific loss $L_{task}$. 

The GATN is complemented by robust sampling using the active learning strategy, $H_{AL}$. Samples are chosen such that they facilitate transfer learning from the pre-trained network as well as trade-off uncertainty for the target network, an indicator of informativeness w.r.t. the target domain \cite{prabhu2020active}. We train the target network through mining samples from the target domain dataset in mini-batches using $H_{AL}$ until the desired budget $B$ is achieved. The steps in the training routine are as follows:
\begin{itemize}
    \item Initialize target network with parameters of the pre-trained source network
    \item Perform the following steps until the total number of samples mined from the target dataset does not exceed $B$.
    \begin{itemize}
        \item Use the AL strategy $H_{AL}$, the pre-trained network and the target network trained in the previous iteration, to mine samples from the target dataset. Accumulate in the set of samples that have already been mined, and annotate them. This gives us a labeled target subset $T_{L}$ and an unlabeled target subset $T_{UL}$.
        \item Jointly train the target network and GATN using $T_{L}$ and $T_{UL}$ by the application of $L_{task}$ and $L_{Tr}$.
    \end{itemize}
\end{itemize}

\subsection{Guided Attention Transfer Network (GATN)}

In this section, we describe the Guided Attention Transfer Network (GATN) in detail. Knowledge transfer at the high-dimensional feature space allows the target network to extract maximal information from the pre-trained network encompassing various aspects of the scene. Hence, we first compute feature maps from the pre-trained network $F_{P}$ and the target network $F_{T}$ for the target dataset input sample $I$. The features $F_{P}$ computed by the pre-trained network on the target network are not well aligned with the intrinsic characteristics of the target domain due to domain shift. Hence, prior to attentive distillation, we perform an initial alignment using a transformation network $\tau$, which is a four-layer fully convolution network. $\tau$ transforms $F_{P}$ to $F_{P-tr}$. The learning of $\tau$, as well as the target feature network, is dictated by a transfer learning loss $L_{Tr}$, which ensures that transformed features $F_{P-tr}$ are well aligned with target domain features $F_{T}$.

Despite the transformation of the features from the pre-trained network, not all knowledge contained in $F_{P-tr}$ is useful to the target domain. This is because $\tau$ is a CNN and has no filtering layers. It is important to transfer only domain agnostic features from $F_{P-tr}$. This is done by the guided attention networks.

$F_{P-tr}$, along with $F_{T}$, are passed to spatial guided attention (GSA) and channel guided attention (GCA) modules. The attention modules compute alignment at the spatial and channel levels for the transformed pre-trained network feature map $F_{P-tr}$ and the target network feature map $F_{T}$. The target network contains finite domain-specific knowledge through the labeled target subset, that can be leveraged to guide transfer learning. Hence, we empower the target network to choose or guide features from the teacher network that it deems suitable for transferring back to itself. We build a guided attention module to do this, which builds on the math synonymous to self-attention~\cite{zhang2019self}. We wish to reflect that the notion of guided attention described in this paper is different from the concept of guided attention in~\cite{li2019guided,mun2017text,pang2019mask}, and is built on the idea described in co-attention~\cite{yu2019deep}. Attention literature~\cite{vaswani2017attention,woo2018cbam,fu2019dual} describes the attention function as mapping a query and a set of key-value pairs to an output, where key, query, value, and output are all vectors. Since target features $F_{T}$ guide the attention process, we designate transformed pre-trained network feature maps as the query vectors. Similarly, key and value are assigned to target network feature maps $F_{P-tr}$. Attention weights are computed as:
\begin{equation}
    A = S(C_{q}(F_{P-tr}^{\top}) \odot C_{k}(F_{T}))^{\top} \odot C_{v}(F_{T}),
\end{equation}
where $\top$, S, and $\odot$ denote transpose operation, softmax and matrix multiplication, respectively. $C_{k}, C_{q}, C_{v}$ denote $1\times 1$ convolutions followed by reshaping of the key, query, and value feature maps respectively. 

We incorporate the notion of guided spatial and channel attentions \cite{woo2018cbam} using spatial and channel level feature vectors \cite{fu2019dual}. The goal of the Guided Spatial Attention (GSA) module (which generates attention representations $A_{GSA}$) is to highlight spatial regions of the transformed pre-trained network feature map, $F_{P-tr}$ that align well with the target feature map, $F_{T}$. The goal of the Guided Channel Attention module (which generates attention representations $A_{GCA}$) is to highlight attributes (or channel level features at each spatial location) of transformed source network features $F_{P-tr}$ at each spatial location that align well with the target network feature maps $F_{T}$. 

\noindent \textbf{Training GATN and Target Network:} 
\label{sec:loss}
We jointly train GATN (consisting of $\tau$ and the guided attention networks), and the target network with the following loss terms:
\begin{itemize}
    \item Transfer learning loss $L_{Tr}$: $F_{P-tr}$ is aligned with $F_{T}$, and the extent to which these feature maps are aligned is delineated by the attention representations. Hence, $L_{Tr}$ is computed as the attention weighted mean squared difference between transformed pre-trained feature maps $F_{P-tr}$ and target network feature maps $F_{T}$. This loss is applied to all target images (labeled as well as unlabeled) and scaled by a hyper-parameter $\lambda_{Tr}$, empirically chosen to $0.1$ or $0.01$ depending on the task. Specifically, we denote $\lambda_{Tr,UL}$ and $\lambda_{Tr,L}$ as the hyper-parameter on unlabeled and labeled target subsets respectively. Mathematically,
    \begin{multline}
        L_{Tr} =  \sum A_{\small{GSA}} * [F_{P-tr}-F_{T}]^2 + \\
        \sum A_{\small{GCA}} * [F_{P-tr}-F_{T}]^2.
    \end{multline}
    
    Attention weighting in $L_{Tr}$ ensures that the learning of target domain features is dictated by domain-agnostic features from the pre-trained network, which are relevant to the target domain.   
    \item Task specific loss $L_{Task}$: To learn target domain-specific information, we apply traditional task losses on the output of target task head, $TH_{T}$, \textit{viz.} multi-class cross entropy for classification and semantic segmentation, and focal loss for object detection, for the labeled subset of target domain images. This loss is scaled by a hyper-parameter $\lambda_{pseudo}$ typically set to $1.0$ \cite{kothandaraman2021domain}.
\end{itemize}
\subsection{Active Learning Strategy $H_{AL}$}
The training of GATN and the target network involves a task specific loss on the labeled subset of the target domain. Hence, the effectiveness of GATN, and robustness and generalizability of the target network depends on the samples mined by the AL strategy $H_{AL}$. Thus, it is important to annotate samples that facilitate transfer learning from the pre-trained network. At the same time, to encode all aspects of target domain-specific knowledge, the target network needs to learn from samples that it is unsure about. Hence, $H_{AL}$ needs to strike a fine balance between transferability from the pre-trained network, and pertinence w.r.t. to the target domain.
\begin{itemize}
    \item \textit{Transferability from the pre-trained network:} The uncertainty of the pre-trained network in computing the outputs of target domain samples is an indicator of the transferability of the corresponding features to the target network. High uncertainty implies low transferability and vice versa. To compute the transferability score, we threshold the final softmax output of the pre-trained network, $N_{P}$, to compute pseudo label map for the target samples. The task-specific loss is computed between the output of the pre-trained network and the pseudo-labels. The transferability score, $A_G$ is the the total $l_2$ norm of the gradient (without any gradient update - since the network is frozen) over the pretrained network \textit{wrt} the computed loss. Low gradient implies high confidence (or low uncertainty) and hence high transferability from the source network.
    \item \textit{Uncertainty of the target network}: The softmax output of the target network provides the class-wise probability score map $\mathbf{p}$. This is used to compute the entropy score $A{E}$ as $-\sum p logp$. High entropy for a target sample indicates high uncertainty and hence should be selected for labeling.
    %\item \textit{Diversity of the target set}: The output feature $F_T$ for target samples for clustered using k-means \cite{ash2019deep}. The mean distance of each target sample from the previously annotated target points gives the diversity score $A_{D}$. 
\end{itemize}
A combination of the above two measures is maximized greedily to mine the samples for labeling:   
\begin{equation}
    H_{AL} = -\lambda_{G} \log A_{G} + \lambda_{E} \log A_{E} %+ \lambda_{D} \log A_{D}, 
\end{equation}
where $\lambda_{G}, \lambda_{E}$, are binary variables ($0/1$) that toggle the metrics used for sampling. For the first batch of AL, we set $\lambda_{G}=1$, and $\lambda_{E}=0$. This is because at the start of training, there is no knowledge available about the target dataset. Thus, we sample images that have good confidence with respect to the pre-trained network for the first round. For the subsequent rounds of sampling, we follow a simple weighting scheme where transferability and uncertainity are given equal importance.

\section{Experiments and Results}
We present results across the classification, detection and segmentation tasks, with standard evaluation metrics. Under classification settings, \model, even without access to annotated source data, performs similar to or better than (with a variance of $0.5$\% in accuracy) prior active domain adaptation methods~\cite{su2020active,prabhu2020active} that use large amounts (more than $100k$ samples) of annotated source data. Furthermore, we conduct experiments on MNIST under $2$ distinct cases of shift in output label space, and show that \model~can achieve atleast $99.4$\% of the accuracy obtained when there is no shift in label space. Thus, \model~can handle shifts in label space. Our experiments on CityScapes for semantic segmentation at various budgets reveal an improvement of atleast $5.57$\% over source-free fine-tuning (\emph{i.e.}  training the model without $L_{tr}$). Finally, we conduct experiments on document layout adaptation from PubLayNet to DSSE where there is a shift in output label space, \model~imparts a relative improvement of $31.3$\% over fine-tuning (\emph{i.e.} training the model without $L_{tr}$). 
\begin{table}
% \footnotesize
\label{tab:mnist_main}
\begin{subtable}{0.9\columnwidth}
\centering
\scriptsize
% \begin{center}
\resizebox{0.9\columnwidth}{!}
{
\begin{tabular}{c c c c c}
\toprule
Method & Source Data & B=$100$ & B=$200$ & B=$300$  \\ 
\toprule
Source only accuracy: $62.25$ \\
\midrule 
O-ALDA \cite{saha2011active} & \cmark & $79.10$ & $81.40$ & $82.70$  \\
%Random \cite{su2020active} & \cmark & $92.90$ & $94.70$ & $95.20$ \\
CDAN \cite{long2017conditional} + Entropy \cite{su2020active} & \cmark & $93.10$ & $94.60$ & $95.00$ \\
CDAN \cite{long2017conditional} + BvSB \cite{su2020active} & \cmark & $94.20$ & $95.00$ & $95.90$ \\
CDAN \cite{long2017conditional} + Uniform  \cite{prabhu2020active} & \cmark & $90.00$ & $94.00$ & $94.50$ \\
CDAN \cite{long2017conditional} + BADGE \cite{ash2019deep} & \cmark & $92.90$ & $94.90$ & $96.50$ \\
SSDA MME \cite{saito2019semi} & \cmark & $93.00$ & $95.00$ & $95.50$ \\
AADA \cite{su2020active} & \cmark & $94.20$ & $95.20$ & $95.50$ \\
CLUE \cite{prabhu2020active} & \cmark & $95.50$ & $96.20$ & $96.50$ \\
\rowcolor{RowColorCode}
\model~ & \xmark & $91.64$ & $95.96$ & $\mathbf{97.16}$ \\
\bottomrule
\end{tabular}
}
\caption{\small{\textbf{Results on adapting from SVHN to MNIST:} With a budget of $300$ images ($0.5\%$ of target data MNIST, last column of table), we show that, even without source data, \model~outperforms prior work on active domain adaptation, using annotated source data. }}
\label{tab:mnist_main}
\end{subtable}
\hspace{10pt}
\begin{subtable}{0.4\textwidth}
% \footnotesize
\centering
% \begin{center}
\resizebox{0.9\textwidth}{!}
{
\begin{tabular}{c c c c c c}
\toprule
Method & $1000$ & $2000$ & $4000$ & $10000$ \\
\toprule
Source only accuracy: $27.27$ \\
\midrule 
FT+Uniform & $68.0$ & $76.2$ & $80.0$ & $84.7$ \\
FT+Entropy & $68.0$ & $75.1$ & $81.2$ & $87.8$ \\
FT+BADGE \cite{ash2019deep} & $70.1$ & $79.2$ & $83.7$ &  $88.1$ \\
FT+Coreset \cite{sener2017active} & $70.0$ & $78.8$ & $82.8$ & $88.2$ \\
FT+Margin \cite{roth2006margin}& $71.0$ & $78.0$ & $83.2$ & $88.4$ \\
FT+CLUE \cite{prabhu2020active}& $72.1$ & $76.4$ & $83.0$ & $87.8$ \\
\rowcolor{RowColorCode}
\model~ & $\mathbf{74.2}$ & $\mathbf{82.2}$ & $\mathbf{86.6}$ & $\mathbf{88.6}$ \\
\bottomrule
\end{tabular}
}
\caption{\small{\textbf{Results on adapting from MNIST to SVHN:} We compare with prior methods on active learning, and demonstrate state-of-the-art performance.}}
\label{tab:svhn_main}
\end{subtable}

% \vspace{-10pt}
\caption{\small{\textbf{Results on digits classification in the shared label space setting }}}
% \end{center}
% \end{center}
\end{table}

\noindent \textbf{Reproducibility:} We provide a detailed description of the datasets, hyperparameters and training details in the supplementary material. We also include the codes for GATN and $H_{AL}$, and links to external repositories to setup the datasets, task-specific training scripts along with detailed training instructions.

\subsection{Image Classification}

We present our results on digits classification datasets under two settings: 
\begin{inparaenum}[(i)]
\item{shared label space, and}
\item{shift in label space.}
\end{inparaenum} In the shared label space setting, the label space that the pre-trained network was trained on and the label space of the target domains are the same. In the label space shift space setting, the target dataset contains labels not used for training the source network. We used Resnet-101 features for the experiment, consistent with the baselines. We set $\lambda_{Tr}=0.01$, and $\lambda_{G}$ to $1.0$  and $\lambda_{E}$ to $1.0$ after first round of sampling. In addition, we generate pseudo-labels from the target network for the unlabeled subset and add it to the $L_{task}$. We use feature heads from the penultimate layer of the ResNet-101 classifier backbone. Due to space constraints, we report results on the VISDA dataset in the supplementary material. 
\begin{table}
% \footnotesize
\begin{subtable}{0.4\textwidth}
\centering
\scriptsize
\resizebox{0.9\textwidth}{!}
{
\begin{tabular}{c c c}
\toprule
Method & Budget & Accuracy  \\ 
\toprule
SDDA (WACV 2021) \cite{kurmi2021domain} & - & $75.5$\\
SDDA-P (WACV 2021) \cite{kurmi2021domain} & - & $76.3$ \\
\model~ & $0.16\%$ & $91.64$ \\
\bottomrule
\end{tabular}
}
\caption{SVHN to MNIST}
\end{subtable}
%\hspace{10pt}
\begin{subtable}{0.4\textwidth}
\centering
\resizebox{0.9\textwidth}{!}
{
\begin{tabular}{c c c}
\toprule
Method & Budget & Accuracy  \\ 
\toprule
SDDA (WACV 2021) \cite{kurmi2021domain} & - & $42.2$\\
SDDA-P (WACV 2021) \cite{kurmi2021domain} & - & $43.6$ \\
\model~ & $0.18\%$ & $74.2$ \\
\bottomrule
\end{tabular}
}
\caption{MNIST to SVHN}
\end{subtable}
\caption{\small{\textbf{Comparing \model~with prior art on SFDA on digits datasets.} Budget reflects the percentage of total number of target samples used for active learning. We demonstrate that our SF-ADA approach outperforms prior art on SFDA by a large margin using a very small proportion of annotated target samples.}}
\label{tab:mnist_sfda_sota}
% \end{center}
\end{table}

\noindent \textbf{Shared label-space setting, SVHN to MNIST:} Table~\ref{tab:mnist_main} contains results of adapting from SVHN to MNIST, at various budgets. Concurrent with our intuition, the accuracy is better at higher budgets. When benchmarked at a budget of $300$ samples, which is $0.5\%$ of the total samples that MNIST contains, we observe that \model, even without any annotated source data, outperforms prior work on active domain adaptation which use large amounts of annotated source data ($600000$ images). Moreover, we observe that the accuracy of $97.16\%$ with $300$ images is $97.64\%$ of fully supervised accuracy with $60000$ images. 

\noindent \textbf{Comparisons with SFDA + AL methods.} \model~outperforms prior work on ADA. Since ADA (or AL + DA) methods are better than AL + SFDA methods, by transitivity, a holistic solution for SF-ADA, such as \model~, is more beneficial than a naive combination of SFDA and AL. 

\noindent \textbf{Shared label-space setting, MNIST to SVHN:} We show results on adaptation from MNIST to SVHN in Table~\ref{tab:svhn_main}. The complexity of SVHN is higher than that of MNIST, reflected in the source-only accuracy which stands at $27\%$. As per the prior work, we cap the net budget for mining samples via active learning at $10000$ images, which is $1.8\%$ of the total size of the dataset. We demonstrate state-of-the-art performance at varying budget of $1000$, $2000$, $4000$ and $10000$ images. Moreover, the accuracy at $10000$ images is $93.44\%$ of fully supervised accuracy, which uses almost $555,555$ images.  
\begin{table}
% \footnotesize
\centering
\scriptsize
% \begin{center}
\resizebox{0.99\columnwidth}{!}
{
\begin{tabular}{c c c c c c c c c c c c}
\toprule
Class/Exp. & mean & $0$ & $1$ & $2$ & $3$ & $4$ & $5$ & $6$ & $7$ & $8$ & $9$  \\ 
\midrule
\multicolumn{12}{c}{Case 1: Remove the digits `3' and `9' from source SVHN} \\
\midrule 
Source only & $56.88$&	$69.20$&	$86.80$&	$79.10$&	\cellcolor{RowColorCode}$0.00$&	$53.80$&	$95.70$&	$41.00$&	$78.20$&	$63.00$	&\cellcolor{RowColorCode}$0.00$\\
B=$100$ &$88.29$&	$97.80$&	$98.90$&	$94.60$&	\cellcolor{RowColorCode}$83.80$&	$94.30$&	$96.60$&	$84.90$&	$91.00$&	$93.80$&	\cellcolor{RowColorCode}$48.20$\\
B=$200$ & $96.27$&	$98.50$&	$98.60$&	$98.10$&	\cellcolor{RowColorCode}$94.30$&	$96.70$&	$97.80$&	$97.90$&	$93.60$&	$92.80$&	\cellcolor{RowColorCode}$94.40$\\
B=$300$ &$96.61$&	$99.10$&	$98.70$&	$98.10$&	\cellcolor{RowColorCode}$95.00$&	$97.50$&	$97.90$&	$98.00$&	$91.30$	&$95.40$	& \cellcolor{RowColorCode}$97.10$\\
\midrule 
\multicolumn{12}{c}{Case 2: Remove the digits `7', `5', `4', `1' from source SVHN} \\
\midrule
Source only & $41.90$&	\cellcolor{RowColorCode}$0.00$&	$70.80$&	$65.60$&	$89.60$&	\cellcolor{RowColorCode}$0.00$&	\cellcolor{RowColorCode}$0.00$&	$83.00$&	\cellcolor{RowColorCode}$0.00$&	$66.00$&	$47.60$\\
B=$100$ & $87.11$&	\cellcolor{RowColorCode}$98.80$&	$99.40$&	$93.80$&	$97.30$&	\cellcolor{RowColorCode}$82.70$&	\cellcolor{RowColorCode}$49.90$&	$94.20$&	\cellcolor{RowColorCode}$85.60$&	$66.40$&	$95.40$\\
B=$200$ & $92.53$&	\cellcolor{RowColorCode}$97.60$&	$98.90$&	$94.10$&	$97.40$&	\cellcolor{RowColorCode}$96.10$&	\cellcolor{RowColorCode}$60.80$&	$95.90$&	\cellcolor{RowColorCode}$90.50$&	$94.50$&	$94.60$\\
B=$300$ & $97.00$&	\cellcolor{RowColorCode}$99.20$&$98.90$&	$97.20$&	$99.10$&	\cellcolor{RowColorCode}$96.20$&	\cellcolor{RowColorCode}$95.70$&	$97.50$&	\cellcolor{RowColorCode}$95.30$	&$94.70$&	$96.30$\\
\bottomrule
\end{tabular}
}
\caption{\small{\textbf{Results on adaptation from SVHN to MNIST, for label space shift.} We consider two scenarios: case 1 - source data does not contain digits `3' and `9', case 2 - source data does not contain digits `7', `5', `4', `1'. \model, with $H_{AL}$ and GATN, achieves $99.4\%$ and $99.8\%$ of the accuracy in the scenario with no label shift. The accuracy at a budget of $300$ images, without label shift, is $97.16\%$}}
\label{tab:mnist_zeroshot}
% \vspace{-10pt}
% \end{center}
\end{table}

\noindent \textbf{Shift in label space, SVHN to MNIST:} In Table~\ref{tab:mnist_zeroshot}, we present results on adapting from SVHN to MNIST under a shift in label space. In case 1, we train the source network on SVHN after removing samples corresponding to two classes (class $3$ and class $9$, randomly chosen). Similarly, in case 2, we remove $4$ classes from the source dataset. Direct testing reveals that the accuracy for these classes is $0$. Sampling with our AL strategy \S\ref{sec:loss} and gradually training with GATN gradually up to a budget of $300$ images (again, $0.5\%$ of target samples) restores accuracy to $96.61\%$ and $97.00\%$, respectively, which is $99.4\%$ and $99.8\%$ of the accuracy achieved in the scenario with no label shift. Thus, our method works well when there is a shift in label space.
\begin{table}
% \footnotesize
\centering
\scriptsize
% \begin{center}
\resizebox{0.8\columnwidth}{!}
{
\begin{tabular}{c c c c c c}
\toprule
\multicolumn{2}{c}{GATN Specification} & \multicolumn{2}{c}{AL Specification} & Budget & Acc \\ 
$\lambda_{Tr,UL}$ & $\lambda_{Tr,UL}$ & $A_{G}$ & $A_{E}$ & & \\
\midrule 
\multicolumn{4}{c}{Baseline source-free accuracy: $62.25\%$}\\
\midrule 
$0$ & $0$  & \cmark & \xmark & $300$ & $88.96$ \\
$0$ & $0.01$ & \cmark & \xmark & $300$ & $89.56$ \\
$0.01$ & $0.01$ & \cmark & \xmark & $300$ & $92.56$ \\
$0.01$ & $0.01$ & \cmark & \cmark & $300$ & $97.16$ \\
\bottomrule
\end{tabular}
}
\caption{\small{Ablation Experiments on adaptation from SVHN to MNIST}}
\label{tab:al_ablations}
%\vspace{-20pt}
% \end{center}
\end{table}

\noindent \textbf{Ablation experiments:} We present ablation experiments on adaptation from SVHN to MNIST in Table \ref{tab:al_ablations}. We set the net active learning budget at $300$ images. Since the network has no prior knowledge about the target domain, the uncertainity metric can be applied only from the second round of active sampling. In the first round of active sampling, we apply only the transferability metric. In the first experiment, we study the impact of training the target network, without GATN, with all samples mined using just the transferability score, the accuracy is $88.96$. This proves that even in the absence of knowledge transfer from the pretrained network, samples mined using the pretrained networks' confidence score is advantageous since the target network is initialized with the weights of the pretrained network. Next, we apply the distillation loss dictated by GATN only on the labeled subset and correspondingly set $L_{L,Tr}=0.01$. We observe that knowledge transfer, in addition to sampling using the transferability score improves performance by $0.6\%$, on an absolute scale. This reinforces the quality of samples mined by the pretrained network. Next, we apply the distillation loss to the unlabeled subset as well ($L_{Tr,UL}$), which leads to an absolute improvement of $3\%$. This is an indicator of GATN's selective knowledge distillation capabilities, where only useful features are distilled to cumulatively improve performance. Finally, we experiment by using the uncertainity score, as well as the diversity score, to achieve an accuracy of $97.16\%$, an absolute improvement of $34.91\%$ over the baseline and an absolute improvement of $5.6\%$ over Experiment 3. Hence, the best knowledge transfer using GATN is obtained when samples are mined intelligently (Experiment 4). 

\noindent \textbf{Ablation experiment on the modulation network $\tau$.} The accuracy on MNIST $\longrightarrow$ SVHN, with $300$ images, without the modulation network is $94.15\%$ while the accuracy with the modulation network is $97.16\%$. 

\begin{table*}
% \footnotesize
\centering
\scriptsize
% \begin{center}
\resizebox{0.99\textwidth}{!}{
\begin{tabular}{c c c c c c c c c c c c c c c c c c c c c c}
\toprule
Experiment & mIoU & mAcc & \rotatebox{90}{Road} & \rotatebox{90}{Sidewalk} & \rotatebox{90}{Building} & \rotatebox{90}{Wall} & \rotatebox{90}{Fence} & \rotatebox{90}{Pole} & \rotatebox{90}{Light} & \rotatebox{90}{Sign} & \rotatebox{90}{Veg} & \rotatebox{90}{Terrain} & \rotatebox{90}{Sky} & \rotatebox{90}{Person} & \rotatebox{90}{Rider} & \rotatebox{90}{Car} & \rotatebox{90}{Truck} & \rotatebox{90}{Bus} & \rotatebox{90}{Train} & \rotatebox{90}{MBike} & \rotatebox{90}{Bike}  \\
\midrule
Source only & $34.91$ &	$77.84$ & $70.14$ &	$21.6$&	$76.27$&	$18.8$&	$16.27$&	$21.31$&	$27.85$&	$15.40$&	$77.67$&	$31.29$&	$74.83$&	$49.47$&	$3.60$&	$79.45$&	$28.71$&	$31.39$&	$4.70$	&$12.43$&	$2.10$\\
TENT \cite{wang2020tent} (Source-free) & $38.9$ & - &$87.3$ &$39.0$ &$79.8$& $24.3$& $19.6$& $21.2$& $25.1$& $16.6$& $83.8$& $34.7$& $77.7$& $57.9$& $17.8$& $85.0$& $24.9$& $20.8$& $2.0$& $16.6$& $4.5$  \\ 
S4T \cite{prabhu2021s4t} (Source-free) & $43.98$ & - & $88.2$ & $44.2$ & $84.4$ & $28.9$ & $27.6$ & $38.6$ & $41.5$ & $8.4$ & $86.3$ & $41.0$ & $79.2$ & $58.7$ & $25.3$ & $85.4$ & $20.1$ & $26.4$ & $6.3$ & $10.8$ & $8.4$    \\
\midrule 
B=$50$ & \textbf{$45.93$}&	\textbf{$89.22$} & $92.09$&	$52.57$&	$83.43$&	$23.72$&	$18.37$&	$33.33$&	$35.90$&	$44.01$&	$84.24$&	$39.23$&	$85.82$&	$55.39$&	$20.16$&	$84.02$&	$38.57$&	$37.77$&	$2.72$&	$16.09$&	$25.26$\\
B=$100$ &\textbf{$50.98$} & \textbf{$90.33$} & $93.6$&	$57.79$&	$84.16$&	$23.4$&	$21.98$&	$36.07$&	$38.12$	&$45.8$&	$85.39$&	$41.33$&	$86.34$&	$57.67$&	$30.41$&	$86.1$	&$43.81$&	$45.02$&	$26.02$&	$19.15$&	$46.48$\\
B=$200$ & \textbf{$53.34$}&	\textbf{$91.18$}& $94.82$&	$63.83$&	$85.29$&	$29.01$&	$27.85$	&$36.84$&	$39.84$&	$47.53$&	$86.33$&	$42.16$&	$88.4$&	$60.16$&	$31.85$&	$86.88$&	$48.64$	&$48.45$&	$26.29$&	$20.46$&	$48.91$\\
B=$500$ & \textbf{$56.59$}&	\textbf{$92.09$} & $95.59$&	$68.77$&	$86.41$&	$33.08$&	$34.88$&	$39.49$&	$42.54$&	$52.44$&	$87.30$&	$48.17$&	$89.73$&	$62.96$&	$33.91$	& $88.18$&	$53.67$&	$52.41$&	$29.08$&	$23.46$	& $53.15$
\\ 
\bottomrule
\end{tabular}
}
\caption{\small{\textbf{GTA5 to CityScapes Adaptation:} We show that \model~ imparts a relative improvement of $31.5\%$, $46.03\%$, $52.5\%$ and $62.1\%$ over the baseline source model with budgets of $50$, $100$, $200$, and $500$ images, with $3\times-25\times$ improvement on specific classes like ``Bike'', ``Train'', ``MBike'', ``Sidewalk'', etc.}}
\label{tab:cityscapes_main}
% \end{center}
% \vspace{-10pt}
%\vspace{-20pt}
\end{table*}

\subsection{Autonomous Driving: Synthetic to Real Segmentation on CityScapes}
We conduct experiments on a dense pixel-level task, segmentation, where we adapt from GTA5 ($25000$ images) to CityScapes. To effectively transfer from GTA5 and address uncertainty of the target network while sampling, we set $\lambda_{G}=\lambda_{E}=1$ from the second round of sampling. %Unlike classification, class-wise k-means clustering cannot be applied to segmentation, hence we set $\lambda_{K}=0$. 
We set $\lambda_{Tr}=0.01$. We use feature heads from the layer 3 of the underlying DeepLabv2 ResNet-101 backbone \cite{chen2017rethinking}. We present the results in Table~\ref{tab:cityscapes_main}. In the first row, we directly test the pre-trained GTA5 model, which gives an mIoU of $34.91$. We next apply $H_{AL}$ for batch active learning, and train the network using GATN after each round of sampling. A cumulative budget of $50$ images, $100$ images, $200$ images, and $500$ images leads to relative improvements (over the source only mIoU) of $31.5\%$, $46.03\%$, $52.5\%$ and $62.1\%$ respectively. Classes like `Sidewalk', `Wall', `Fence', `Pole', `Sign', `Rider', `Train', `MBike', `Bike' have very low mIoU ($\sim20$\% or less) when directly transferred from the source model. We show that \model~improves performance by $3\times-25\times$. Classwise comparisons with prior art on SFDA reveal that our AL heuristic $H_{AL}$ strategically chooses classes like Bike and MBike with low confidence or high uncertainity to boost performance, while not compromising on performance w.r.t. transferable classes like road and sky. 

\begin{table}
% \footnotesize
\centering
\scriptsize
% \begin{center}
\resizebox{0.4\textwidth}{!}
{
\begin{tabular}{c c}
\toprule
Method & mIoU  \\ 
\midrule
UBNA \cite{klingner2022unsupervised} (WACVW 2022) & $36.1$ \\
UBNA+ \cite{klingner2022unsupervised} (WACVW 2022) & $36.5$ \\
TENT \cite{wang2020tent} (ICLR 2021) & $38.86$ \\
TENT + MS \cite{wang2020tent} (ICLR 2021) & $36.89$ \\
SFDA (w/o IPSM) \cite{liu2021source} (CVPR 2021) & $41.35$ \\
SFDA \cite{liu2021source} (CVPR 2021) & $43.16$ \\
URMA \cite{fleuret2021uncertainty} (CVPR 2021) & $45.1$ \\
S4T \cite{prabhu2021s4t} (ArXiv 2021) & $43.98$ \\
S4T + MS \cite{prabhu2021s4t} (ArXiv 2021) & $44.83$ \\
\rowcolor{RowColorCode}
\textbf{\model~} & \textbf{$45.93$} \\
\bottomrule
\end{tabular}
}
\caption{\small{\textbf{Comparisons with the state-of-the-art SFDA approaches for adaptation from GTA to CityScapes.} We show that \model~achieves state-of-the-art mIoU with a small budget of $50$ images, the mIoU obtained by naive finetuning \cite{wang2020alleviating} on $50$ images is $39.5$.}}
\label{tab:cityscapes_sota}
%\vspace{-20pt}
\end{table}

\noindent \textbf{Comparisons with SFDA methods:} Table~\ref{tab:cityscapes_sota} shows the comparison of \model~ with prior SFDA methods. \model~ performs better with a budget of only 50 images ($1.5 \%$).
%Semi-supervised domain adaptation methods \cite{wang2020alleviating,huang2021semi,chen2021semi} use data augmentation techniques as well as source data while training, and hence are not pertinent to SF-ADA.

%In the third part of the table, we compare against the current state-of-the-art in semi-supervised domain adaptation, which uses large amounts of annotated source data, as well a subset of labeled target data. At the same budget of labeled target data and without any source data, we show that \model~outperforms the state-of-the-art semi-supervised DA method.
\begin{table}
% \footnotesize
\begin{subtable}{0.45\columnwidth}
\centering
\scriptsize
\resizebox{0.9\textwidth}{!}
{
\begin{tabular}{c c c}
\toprule
Budget $B$ & ($\lambda_{Tr,L},$ & $\lambda_{Tr,UL}$)  \\ & (0.1,0) & (0.1,0.1)\\ 
\toprule
$50$ & $44.62$ & $45.93$\\
$100$ & $48.72$ & $50.98$ \\
$200$ & $50.49$ & $53.34$\\
$500$ & $53.46$ & $56.59$\\
\bottomrule
\end{tabular}
}
\caption{\small{}}
\end{subtable}
%\hspace{10pt}
\begin{subtable}{0.3\columnwidth}
\centering
\resizebox{0.9\textwidth}{!}
{
\begin{tabular}{c c c}
\toprule 
GSA & GCA & mIoU \\
\midrule
\xmark & \xmark & $39.50$ \\
\cmark & \xmark & $45.43$ \\
\xmark & \cmark & $45.45$ \\
\cmark & \cmark & $45.93$ \\
\bottomrule
\end{tabular}
}
\caption{\small{}}
\end{subtable}
%\hspace{10pt}
\begin{subtable}{0.2\columnwidth}
\centering
\resizebox{0.9\textwidth}{!}
{
\begin{tabular}{c c}
\toprule 
$H_{AL}$ & mIoU \\
\midrule
$A_{G}$ & $50.29$ \\
$A_{G}$ + $A_{E}$ & $56.59$ \\
\bottomrule
\end{tabular}
}
\caption{\small{}}
\end{subtable}
\caption{\small{\textbf{Ablation experiments for adaptation from GTA5 to CityScapes.} In table (a), we study ablations on transfer loss for GATN, Table (b) shows ablations on GATN components at a budget of $50$ images. In table (c), we ablate on the active learning heuristic, $H_{AL}$ at a budget of $500$ images.}}
\label{tab:cityscapes_ablations}
%\vspace{-20pt}
% \end{center}
\end{table}

\begin{table}
% \footnotesize
\centering
\scriptsize
% \begin{center}
\resizebox{0.99\columnwidth}{!}
{
\begin{tabular}{c c c c c c c c}
\toprule
Experiment & mAP & \rotatebox{90}{Text} & \rotatebox{90}{Caption} & \rotatebox{90}{Figure} & \rotatebox{90}{Table} & \rotatebox{90}{List} & \rotatebox{90}{Section}   \\
\midrule
FT w/o \model~ & $23.11$ & $36.12$ & $13.49$ & $25.60$ & $22.24$ & $29.57$ & $11.64$ \\
\rowcolor{RowColorCode}
FT w. \model~ & $30.36$ & $44.59$ & $11.61$ & $35.57$ & $24.80$ & $37.48$ & $28.18$ \\
\bottomrule
\end{tabular}
}
\caption{\small{\textbf{Adaptation for document layout detection from PubLayNet to DSSE:} Fine-tuning with \model~ improves performance by $31.3\%$ over fine-tuning without \model.}}
\label{tab:dsse}
% \end{center}
\end{table}

\noindent \textbf{Ablation experiments:} We present ablation experiments in Table~\ref{tab:cityscapes_ablations}. In Table~\ref{tab:cityscapes_ablations}(a), we study the effectiveness of GATN at various budgets. In the second column, we apply the transfer learning loss dictated by GATN to only the labeled subset, along with active learning using $H_{AL}$. The third column reflects mIoUs obtained by using our complete model, applying the transfer learning on labeled as well as unlabeled subsets along with active learning using $H_{AL}$. A comparison of the second and third columns indicates the benefits of selective transfer learning using GATN. Moreover, our model results in relative improvements of $16.27\%$, $16.92\%$, $13.24\%$ and $5.57\%$ over the baseline numbers \cite{wang2020alleviating} obtained by naively finetuning the target network (with pre-trained weights initialization) without GATN, and by random sampling \cite{wang2020alleviating}, at budgets of $50$ images, $100$ images, $200$ images, and $500$ images respectively. In Table~\ref{tab:cityscapes_ablations}(b), we study the effectiveness of the different components of GATN, at a budget of $50$ images. In the first experiment, we do not use either GCA or GSA \cite{wang2020alleviating}. Without GATN, our system will reduce to simple
fine-tuning of the target network with the annotated target
samples. GATN forms a bridge between the pre-trained network and the target network, and removing it would break the knowledge transfer process. In the subsequent experiments, we show the impact of using channel and spatial features. In Table~\ref{tab:cityscapes_ablations}(c), we demonstrate the effectiveness of using the fusion AL heuristic comprising of the transferability score w.r.t. the pretrained network as well as the uncertainity score w.r.t. the target network, as opposed to using just the transferability score. 

\subsection{Document Layout Detection: DSSE}
In this section, we adapt from the medical documents dataset PubLayNet to documents belonging to DSSE, a dataset containing magazines, receipts and posters. The documents in the two domains are quite different. Medical documents are written in a two-column format, with uniform text, figures and tables. In contrast, the target domain, DSSE, a new unseen dataset, is small (only 150 documents) and is extremely diverse. Moreover, PubLayNet has $5$ classes, while DSSE has $6$ classes. Hence, there is a \textit{shift} in label space. Direct testing of PubLayNet on DSSE results in a mAP of $15.67$. Since the dataset is very small, we do not apply $H_{AL}$, and instead directly use all $150$ images for GATN. We use feature heads from the FPN of the underlying RetineNet ResNet-101 backbone \cite{lin2017focal}. Fine-tuning without \model~results in a mAP of $23.11$, and fine-tuning with \model~improves performance by $31.3\%$ to $30.36$.

\section{Conclusions, Limitations and Future Work}

We propose a generic source-free method, \model, for the task of adapting from a pretrained network to target domain, under a possible shift in label space, with the provision to annotate a small budget of samples in the target domain. \model~consists of two complementary components: an active learning strategy $H_{AL}$, and GATN for effective adaptation and sampling. We evaluate the performance across $3$ tasks and show improved or on par performance with methods using source data. One drawback of our method is that we use binary weights for the scores in $H_{AL}$, using learnable weights could be an interesting direction for future work. Moreover, we expect  that \model~can be extended to tasks and modalities where traditional domain adaptation has been useful. These include text classification, neural machine translation, sentiment analysis, cross lingual question answering, and domain stylization. 
%Supplementary: More CS examples; AL samples
%CityScapes AL budget ablations mention in text

\noindent \textbf{Acknowledgements:} This research has been supported by ARO Grants W911NF2110026 and Army Cooperative Agreement W911NF2120076   

\section*{A.1. Datasets}

In this section, we describe the datasets used in our experiments.

\subsection{Classification}
\begin{itemize}
    \item MNIST \cite{deng2012mnist}: MNIST is a handwritten digits dataset, with $60,000$ samples for training and $10,000$ samples for testing. It can be downloaded at http://yann.lecun.com/exdb/mnist/.
    \item SVHN \cite{sermanet2012convolutional}: SVHN is a house street numbers dataset, and has cropped digits with character wise ground-truth in MNIST format. It has over $600,000$ images and is a much more realistic dataset than MNIST. It can be downloaded at http://ufldl.stanford.edu/housenumbers/.
    \item VISDA-17 \cite{peng2017visda}: VISDA is dataset designed for synthetic to real adaptation. The synthetic images are 2D renderings of 3D models generated from various angles and lighting conditions. The real images correspond to natural scene objects. It can be downloaded at http://ai.bu.edu/visda-2017/.
\end{itemize}
\subsection{Segmentation}
\begin{itemize}
    \item GTA5 \cite{richter2016playing}: GTA5 is a synthetic driving dataset extracted from the computer game Grand Theft Auto. It has $25000$ high resolution images. The dataset is available at https://download.visinf.tu-darmstadt.de/data/from\_games/.It has $19$ classes compatible with CityScapes. 
    \item CityScapes \cite{cordts2016cityscapes}: CityScapes is a real driving dataset collected in Europe. It has $2975$ high resolution images for training, and $500$ images for testing. The dataset is available for download at https://www.cityscapes-dataset.com/. It has $19$ classes.
\end{itemize}
\subsection{Document Layout Detection}
\begin{itemize}
    \item PubLayNet \cite{zhong2019publaynet}: PubLayNet is a large-scale medical documents dataset consisting of images of pages extracted from scientific medical papers. Medical documents are written in a two-column format, with uniform text, figures and tables. It has $360,000$ images and $5$ classes. The dataset can be downloaded at https://github.com/ibm-aur-nlp/PubLayNet. 
    \item DSSE \cite{yang2017learning}: DSSE contains images of pages extracted from magazines, receipts and posters. DSSE is a small dataset with just $150$ documents, and has $6$ classes, paving way for open-set adaptation from DSSE. The dataset can be downloaded at http://personal.psu.edu/xuy111/projects/cvpr2017\_doc.html.
\end{itemize}

\begin{table}
% \footnotesize
\caption{\textbf{Synthetic to Real Classification on VISDA:} Our source-free method is on par with state-of-the-art methods that use abundant annotated source data (more than 100k samples).}

\centering
% \begin{center}
\resizebox{0.4\columnwidth}{!}{
\begin{tabular}{c c c c}
\toprule
Method & Source Data & B=$10\%$ & B=$20\%$  \\ 
\toprule
Random & \cmark & $82.1$ & $87.2$ \\
UCN \cite{joshi2012scalable}& \cmark & $85.4$ & $90.3$ \\
QBC \cite{nguyen2004active}& \cmark & $84.1$ & $89.6$ \\
Cluster \cite{dagan1995committee}& \cmark & $83.5$ & $89.6$ \\
AADA \cite{su2020active} & \cmark & $84.6$ & $89.7$ \\
ADMA \cite{huang2018cost} & \cmark & $84.8$ & $90.0$ \\
\rowcolor{RowColorCode}
\model~ & \xmark & $84.8$ & $89.3$ \\
\bottomrule
\end{tabular}
}
\label{tab:visda}
% \vpsace
\vspace{-10pt}

% \end{center}
\end{table}

\section*{A.2. Synthetic to Real VISDA17 Classification}
We conduct experiments on the popular VISDA17 dataset for synthetic to real adaptation. For effective transfer, and to address uncertainty of the target network while sampling, we set $\lambda_{G}=\lambda_{E}=1$ from the second round of sampling. VISDA is a huge dataset with a large variety of samples. Hence, we factor in diversity. The output feature $F_T$ for target samples for clustered using k-means \cite{ash2019deep}. The mean distance of each target sample from the previously annotated target points gives the diversity score $A_{D}$. We set the hyperparameter for diversity score $\lambda_{K}=1$ from the second round of sampling. On budgets of $10\%$, and $20\%$ of the total target samples, we achieve accuracies of $84.8\%$ and $89.3\%$ respectively. Though \model~does not use any annotated source data, it achieves accuracies on par with prior work using abundant annotated source data (more than $100k$ samples). 

\section*{A.3. Implementation details}
\paragraph{Hyperparameters:} The transformation network $\tau$ is a four layer convolutional neural network with kernel size 3, and dilation and padding set to $1$. The weight hyperparameter for $L_{Tr}$ is set to $0.1$. Our classification models are trained using 1 GPU with 16GB memory. Our document layout detection and segmentation models are trained using 8 GPUs with 16 GB memory each. All our codes are written using the PyTorch framework. For CityScapes, all images are downsampled by a factor of $2$ using bilinear downsampling. Ground truth maps are downsampled by nearest neighbour downsampling. We retain the input image size for our classification experiments. For document layout detection, we resize the images (and appropriately scale the bounding box coordinates) such that the length of the largest size does not exceed 500. 

\paragraph{Codes:} In the interest of reproducibility, we release the codes for GATN, including the code for the transformation network, the guided attention modules, and their incorporation within DeepLabv2 for segmentation. We release the train and eval scripts for segmentation as well. We also provide the scripts for $H_{AL}$ with the supplementary zip file. We will make these scripts publicly available upon acceptance of the paper.

We also provide the links to public repositories that we used in our experiments for running \model~experiments.

\begin{itemize}
    \item Classification
    \begin{itemize}
        \item Backbone network: https://github.com/tim-learn/SHOT/blob/master/object/network.py
        Please follow the procedure in the deeplab\_multi.py script in the supplementary zip file to incorporate GATN within the classification backbone.
        \item MNIST and SVHN dataloader: https://github.com/tim-learn/SHOT/tree/master/digit/data\_load
        \item VISDA dataloader: https://github.com/VisionLearningGroup/taskcv-2017-public
        \item Training and eval scripts: Please modify the train and eval script in the supplementary zip file to modify code for classification.
    \end{itemize}
    \item Document Layout Detection
    \begin{itemize}
        \item Backbone network, train and test scripts: https://github.com/yhenon/pytorch-retinanet
        \item PubLayNet dataloader: https://github.com/phamquiluan/PubLayNet/\\blob/master/training\_code/datasets/publaynet.py
        \item DSSE dataloader: Use the PubLayNet dataloader to modify.
    \end{itemize}
    \item Semantic Segmentation
    \begin{itemize}
        \item Backbone network, train and test scripts: Please check the supplementary zip file.
        \item GTA5 and CityScapes dataloaders: https://github.com/wasidennis/\\AdaptSegNet/tree/master/dataset
    \end{itemize}
\end{itemize}

{\small
\bibliographystyle{ieee_fullname}
\bibliography{references}

\begin{thebibliography}{10}\itemsep=-1pt

\bibitem{agarwal2022unsupervised}
Peshal Agarwal, Danda~Pani Paudel, Jan-Nico Zaech, and Luc Van~Gool.
\newblock Unsupervised robust domain adaptation without source data.
\newblock In {\em Proceedings of the IEEE/CVF Winter Conference on Applications
  of Computer Vision}, pages 2009--2018, 2022.

\bibitem{ash2019deep}
Jordan~T Ash, Chicheng Zhang, Akshay Krishnamurthy, John Langford, and Alekh
  Agarwal.
\newblock Deep batch active learning by diverse, uncertain gradient lower
  bounds.
\newblock {\em arXiv preprint arXiv:1906.03671}, 2019.

\bibitem{bousmalis2016domain}
Konstantinos Bousmalis, George Trigeorgis, Nathan Silberman, Dilip Krishnan,
  and Dumitru Erhan.
\newblock Domain separation networks.
\newblock {\em Advances in neural information processing systems}, 29:343--351,
  2016.

\bibitem{bouvier2020stochastic}
Victor Bouvier, Philippe Very, Cl{\'e}ment Chastagnol, Myriam Tami, and
  C{\'e}line Hudelot.
\newblock Stochastic adversarial gradient embedding for active domain
  adaptation.
\newblock {\em arXiv preprint arXiv:2012.01843}, 2020.

\bibitem{chen2017rethinking}
Liang-Chieh Chen, George Papandreou, Florian Schroff, and Hartwig Adam.
\newblock Rethinking atrous convolution for semantic image segmentation.
\newblock {\em arXiv preprint arXiv:1706.05587}, 2017.

\bibitem{cordts2016cityscapes}
Marius Cordts, Mohamed Omran, Sebastian Ramos, Timo Rehfeld, Markus Enzweiler,
  Rodrigo Benenson, Uwe Franke, Stefan Roth, and Bernt Schiele.
\newblock The cityscapes dataset for semantic urban scene understanding.
\newblock In {\em Proceedings of the IEEE conference on computer vision and
  pattern recognition}, pages 3213--3223, 2016.

\bibitem{dagan1995committee}
Ido Dagan and Sean~P Engelson.
\newblock Committee-based sampling for training probabilistic classifiers.
\newblock In {\em Machine Learning Proceedings 1995}, pages 150--157. Elsevier,
  1995.

\bibitem{deng2012mnist}
Li Deng.
\newblock The mnist database of handwritten digit images for machine learning
  research [best of the web].
\newblock {\em IEEE Signal Processing Magazine}, 29(6):141--142, 2012.

\bibitem{fleuret2021uncertainty}
Francois Fleuret et~al.
\newblock Uncertainty reduction for model adaptation in semantic segmentation.
\newblock In {\em Proceedings of the IEEE/CVF Conference on Computer Vision and
  Pattern Recognition}, pages 9613--9623, 2021.

\bibitem{fu2021transferable}
Bo Fu, Zhangjie Cao, Jianmin Wang, and Mingsheng Long.
\newblock Transferable query selection for active domain adaptation.
\newblock In {\em Proceedings of the IEEE/CVF Conference on Computer Vision and
  Pattern Recognition}, pages 7272--7281, 2021.

\bibitem{fu2019dual}
Jun Fu, Jing Liu, Haijie Tian, Yong Li, Yongjun Bao, Zhiwei Fang, and Hanqing
  Lu.
\newblock Dual attention network for scene segmentation.
\newblock In {\em Proceedings of the IEEE/CVF Conference on Computer Vision and
  Pattern Recognition}, pages 3146--3154, 2019.

\bibitem{ghifary2016deep}
Muhammad Ghifary, W~Bastiaan Kleijn, Mengjie Zhang, David Balduzzi, and Wen Li.
\newblock Deep reconstruction-classification networks for unsupervised domain
  adaptation.
\newblock In {\em European conference on computer vision}, pages 597--613.
  Springer, 2016.

\bibitem{hong2018conditional}
Weixiang Hong, Zhenzhen Wang, Ming Yang, and Junsong Yuan.
\newblock Conditional generative adversarial network for structured domain
  adaptation.
\newblock In {\em Proceedings of the IEEE Conference on Computer Vision and
  Pattern Recognition}, pages 1335--1344, 2018.

\bibitem{huang2018cost}
Sheng-Jun Huang, Jia-Wei Zhao, and Zhao-Yang Liu.
\newblock Cost-effective training of deep cnns with active model adaptation.
\newblock In {\em Proceedings of the 24th ACM SIGKDD International Conference
  on Knowledge Discovery \& Data Mining}, pages 1580--1588, 2018.

\bibitem{ishii2021source}
Masato Ishii and Masashi Sugiyama.
\newblock Source-free domain adaptation via distributional alignment by
  matching batch normalization statistics.
\newblock {\em arXiv preprint arXiv:2101.10842}, 2021.

\bibitem{joshi2012scalable}
Ajay~J Joshi, Fatih Porikli, and Nikolaos~P Papanikolopoulos.
\newblock Scalable active learning for multiclass image classification.
\newblock {\em IEEE transactions on pattern analysis and machine intelligence},
  34(11):2259--2273, 2012.

\bibitem{kang2019contrastive}
Guoliang Kang, Lu Jiang, Yi Yang, and Alexander~G Hauptmann.
\newblock Contrastive adaptation network for unsupervised domain adaptation.
\newblock In {\em Proceedings of the IEEE/CVF Conference on Computer Vision and
  Pattern Recognition}, pages 4893--4902, 2019.

\bibitem{kim2020domain}
Youngeun Kim, Donghyeon Cho, Kyeongtak Han, Priyadarshini Panda, and Sungeun
  Hong.
\newblock Domain adaptation without source data.
\newblock {\em arXiv preprint arXiv:2007.01524}, 2020.

\bibitem{klingner2022unsupervised}
Marvin Klingner, Jan-Aike Term{\"o}hlen, Jacob Ritterbach, and Tim Fingscheidt.
\newblock Unsupervised batchnorm adaptation (ubna): A domain adaptation method
  for semantic segmentation without using source domain representations.
\newblock In {\em Proceedings of the IEEE/CVF Winter Conference on Applications
  of Computer Vision}, pages 210--220, 2022.

\bibitem{kothandaraman2020ss}
Divya Kothandaraman, Rohan Chandra, and Dinesh Manocha.
\newblock Ss-sfda: Self-supervised source-free domain adaptation for road
  segmentation in hazardous environments.
\newblock {\em arXiv preprint arXiv:2012.08939}, 2020.

\bibitem{kothandaraman2021domain}
Divya Kothandaraman, Athira~M Nambiar, and Anurag Mittal.
\newblock Domain adaptive knowledge distillation for driving scene semantic
  segmentation.
\newblock In {\em WACV (Workshops)}, pages 134--143, 2021.

\bibitem{kundu2020universal}
Jogendra~Nath Kundu, Naveen Venkat, R~Venkatesh Babu, et~al.
\newblock Universal source-free domain adaptation.
\newblock In {\em Proceedings of the IEEE/CVF Conference on Computer Vision and
  Pattern Recognition}, pages 4544--4553, 2020.

\bibitem{kurmi2021domain}
Vinod~K Kurmi, Venkatesh~K Subramanian, and Vinay~P Namboodiri.
\newblock Domain impression: A source data free domain adaptation method.
\newblock In {\em Proceedings of the IEEE/CVF Winter Conference on Applications
  of Computer Vision}, pages 615--625, 2021.

\bibitem{li2020cross}
Kai Li, Curtis Wigington, Chris Tensmeyer, Handong Zhao, Nikolaos Barmpalios,
  Vlad~I Morariu, Varun Manjunatha, Tong Sun, and Yun Fu.
\newblock Cross-domain document object detection: Benchmark suite and method.
\newblock In {\em Proceedings of the IEEE/CVF Conference on Computer Vision and
  Pattern Recognition}, pages 12915--12924, 2020.

\bibitem{li2019guided}
Kunpeng Li, Ziyan Wu, Kuan-Chuan Peng, Jan Ernst, and Yun Fu.
\newblock Guided attention inference network.
\newblock {\em IEEE transactions on pattern analysis and machine intelligence},
  42(12):2996--3010, 2019.

\bibitem{li2020model}
Rui Li, Qianfen Jiao, Wenming Cao, Hau-San Wong, and Si Wu.
\newblock Model adaptation: Unsupervised domain adaptation without source data.
\newblock In {\em Proceedings of the IEEE/CVF Conference on Computer Vision and
  Pattern Recognition}, pages 9641--9650, 2020.

\bibitem{li2018adaptive}
Yanghao Li, Naiyan Wang, Jianping Shi, Xiaodi Hou, and Jiaying Liu.
\newblock Adaptive batch normalization for practical domain adaptation.
\newblock {\em Pattern Recognition}, 80:109--117, 2018.

\bibitem{liang2021source}
Jian Liang, Dapeng Hu, Yunbo Wang, Ran He, and Jiashi Feng.
\newblock Source data-absent unsupervised domain adaptation through hypothesis
  transfer and labeling transfer.
\newblock {\em IEEE Transactions on Pattern Analysis and Machine Intelligence},
  2021.

\bibitem{lin2017focal}
Tsung-Yi Lin, Priya Goyal, Ross Girshick, Kaiming He, and Piotr Doll{\'a}r.
\newblock Focal loss for dense object detection.
\newblock In {\em Proceedings of the IEEE international conference on computer
  vision}, pages 2980--2988, 2017.

\bibitem{liu2021source}
Yuang Liu, Wei Zhang, and Jun Wang.
\newblock Source-free domain adaptation for semantic segmentation.
\newblock In {\em Proceedings of the IEEE/CVF Conference on Computer Vision and
  Pattern Recognition}, pages 1215--1224, 2021.

\bibitem{long2015learning}
Mingsheng Long, Yue Cao, Jianmin Wang, and Michael Jordan.
\newblock Learning transferable features with deep adaptation networks.
\newblock In {\em International conference on machine learning}, pages 97--105.
  PMLR, 2015.

\bibitem{long2017conditional}
Mingsheng Long, Zhangjie Cao, Jianmin Wang, and Michael~I Jordan.
\newblock Conditional adversarial domain adaptation.
\newblock {\em arXiv preprint arXiv:1705.10667}, 2017.

\bibitem{mao2018unpaired}
Xudong Mao and Qing Li.
\newblock Unpaired multi-domain image generation via regularized conditional
  gans.
\newblock {\em arXiv preprint arXiv:1805.02456}, 2018.

\bibitem{maria2017autodial}
Fabio Maria~Carlucci, Lorenzo Porzi, Barbara Caputo, Elisa Ricci, and Samuel
  Rota~Bulo.
\newblock Autodial: Automatic domain alignment layers.
\newblock In {\em Proceedings of the IEEE International Conference on Computer
  Vision}, pages 5067--5075, 2017.

\bibitem{mun2017text}
Jonghwan Mun, Minsu Cho, and Bohyung Han.
\newblock Text-guided attention model for image captioning.
\newblock In {\em Proceedings of the AAAI Conference on Artificial
  Intelligence}, volume~31, 2017.

\bibitem{nguyen2004active}
Hieu~T Nguyen and Arnold Smeulders.
\newblock Active learning using pre-clustering.
\newblock In {\em Proceedings of the twenty-first international conference on
  Machine learning}, page~79, 2004.

\bibitem{pang2019mask}
Yanwei Pang, Jin Xie, Muhammad~Haris Khan, Rao~Muhammad Anwer, Fahad~Shahbaz
  Khan, and Ling Shao.
\newblock Mask-guided attention network for occluded pedestrian detection.
\newblock In {\em Proceedings of the IEEE/CVF International Conference on
  Computer Vision}, pages 4967--4975, 2019.

\bibitem{peng2017visda}
Xingchao Peng, Ben Usman, Neela Kaushik, Judy Hoffman, Dequan Wang, and Kate
  Saenko.
\newblock Visda: The visual domain adaptation challenge.
\newblock {\em arXiv preprint arXiv:1710.06924}, 2017.

\bibitem{prabhu2020active}
Viraj Prabhu, Arjun Chandrasekaran, Kate Saenko, and Judy Hoffman.
\newblock Active domain adaptation via clustering uncertainty-weighted
  embeddings.
\newblock {\em arXiv preprint arXiv:2010.08666}, 2020.

\bibitem{prabhu2021s4t}
Viraj Prabhu, Shivam Khare, Deeksha Kartik, and Judy Hoffman.
\newblock S4t: Source-free domain adaptation for semantic segmentation via
  self-supervised selective self-training.
\newblock {\em arXiv preprint arXiv:2107.10140}, 2021.

\bibitem{rai2010domain}
Piyush Rai, Avishek Saha, Hal Daum{\'e}~III, and Suresh Venkatasubramanian.
\newblock Domain adaptation meets active learning.
\newblock In {\em Proceedings of the NAACL HLT 2010 Workshop on Active Learning
  for Natural Language Processing}, pages 27--32, 2010.

\bibitem{richter2016playing}
Stephan~R Richter, Vibhav Vineet, Stefan Roth, and Vladlen Koltun.
\newblock Playing for data: Ground truth from computer games.
\newblock In {\em European conference on computer vision}, pages 102--118.
  Springer, 2016.

\bibitem{roth2006margin}
Dan Roth and Kevin Small.
\newblock Margin-based active learning for structured output spaces.
\newblock In {\em European Conference on Machine Learning}, pages 413--424.
  Springer, 2006.

\bibitem{rusticus2019document}
Diede Rusticus, Lutz Goldmann, Matthias Reisser, and Mauricio Villegas.
\newblock Document domain adaptation with generative adversarial networks.
\newblock In {\em 2019 International Conference on Document Analysis and
  Recognition (ICDAR)}, pages 1432--1437. IEEE, 2019.

\bibitem{saha2011active}
Avishek Saha, Piyush Rai, Hal Daum{\'e}, Suresh Venkatasubramanian, and Scott~L
  DuVall.
\newblock Active supervised domain adaptation.
\newblock In {\em Joint European Conference on Machine Learning and Knowledge
  Discovery in Databases}, pages 97--112. Springer, 2011.

\bibitem{saito2019semi}
Kuniaki Saito, Donghyun Kim, Stan Sclaroff, Trevor Darrell, and Kate Saenko.
\newblock Semi-supervised domain adaptation via minimax entropy.
\newblock In {\em Proceedings of the IEEE/CVF International Conference on
  Computer Vision}, pages 8050--8058, 2019.

\bibitem{sankaranarayanan2018generate}
Swami Sankaranarayanan, Yogesh Balaji, Carlos~D Castillo, and Rama Chellappa.
\newblock Generate to adapt: Aligning domains using generative adversarial
  networks.
\newblock In {\em Proceedings of the IEEE Conference on Computer Vision and
  Pattern Recognition}, pages 8503--8512, 2018.

\bibitem{sener2017active}
Ozan Sener and Silvio Savarese.
\newblock Active learning for convolutional neural networks: A core-set
  approach.
\newblock {\em arXiv preprint arXiv:1708.00489}, 2017.

\bibitem{sermanet2012convolutional}
Pierre Sermanet, Soumith Chintala, and Yann LeCun.
\newblock Convolutional neural networks applied to house numbers digit
  classification.
\newblock In {\em Proceedings of the 21st International Conference on Pattern
  Recognition (ICPR2012)}, pages 3288--3291. IEEE, 2012.

\bibitem{shen2018wasserstein}
Jian Shen, Yanru Qu, Weinan Zhang, and Yong Yu.
\newblock Wasserstein distance guided representation learning for domain
  adaptation.
\newblock In {\em Thirty-Second AAAI Conference on Artificial Intelligence},
  2018.

\bibitem{sinha2019variational}
Samarth Sinha, Sayna Ebrahimi, and Trevor Darrell.
\newblock Variational adversarial active learning.
\newblock In {\em Proceedings of the IEEE/CVF International Conference on
  Computer Vision}, pages 5972--5981, 2019.

\bibitem{su2020active}
Jong-Chyi Su, Yi-Hsuan Tsai, Kihyuk Sohn, Buyu Liu, Subhransu Maji, and
  Manmohan Chandraker.
\newblock Active adversarial domain adaptation.
\newblock In {\em Proceedings of the IEEE/CVF Winter Conference on Applications
  of Computer Vision}, pages 739--748, 2020.

\bibitem{tsai2018learning}
Yi-Hsuan Tsai, Wei-Chih Hung, Samuel Schulter, Kihyuk Sohn, Ming-Hsuan Yang,
  and Manmohan Chandraker.
\newblock Learning to adapt structured output space for semantic segmentation.
\newblock In {\em Proceedings of the IEEE conference on computer vision and
  pattern recognition}, pages 7472--7481, 2018.

\bibitem{tzeng2017adversarial}
Eric Tzeng, Judy Hoffman, Kate Saenko, and Trevor Darrell.
\newblock Adversarial discriminative domain adaptation.
\newblock In {\em Proceedings of the IEEE conference on computer vision and
  pattern recognition}, pages 7167--7176, 2017.

\bibitem{vaswani2017attention}
Ashish Vaswani, Noam Shazeer, Niki Parmar, Jakob Uszkoreit, Llion Jones,
  Aidan~N Gomez, {\L}ukasz Kaiser, and Illia Polosukhin.
\newblock Attention is all you need.
\newblock In {\em Advances in neural information processing systems}, pages
  5998--6008, 2017.

\bibitem{vu2019advent}
Tuan-Hung Vu, Himalaya Jain, Maxime Bucher, Matthieu Cord, and Patrick
  P{\'e}rez.
\newblock Advent: Adversarial entropy minimization for domain adaptation in
  semantic segmentation.
\newblock In {\em Proceedings of the IEEE/CVF Conference on Computer Vision and
  Pattern Recognition}, pages 2517--2526, 2019.

\bibitem{wang2014new}
Dan Wang and Yi Shang.
\newblock A new active labeling method for deep learning.
\newblock In {\em 2014 International joint conference on neural networks
  (IJCNN)}, pages 112--119. IEEE, 2014.

\bibitem{wang2020tent}
Dequan Wang, Evan Shelhamer, Shaoteng Liu, Bruno Olshausen, and Trevor Darrell.
\newblock Tent: Fully test-time adaptation by entropy minimization.
\newblock {\em arXiv preprint arXiv:2006.10726}, 2020.

\bibitem{wang2020alleviating}
Zhonghao Wang, Yunchao Wei, Rogerio Feris, Jinjun Xiong, Wen-Mei Hwu, Thomas~S
  Huang, and Honghui Shi.
\newblock Alleviating semantic-level shift: A semi-supervised domain adaptation
  method for semantic segmentation.
\newblock In {\em Proceedings of the IEEE/CVF Conference on Computer Vision and
  Pattern Recognition Workshops}, pages 936--937, 2020.

\bibitem{woo2018cbam}
Sanghyun Woo, Jongchan Park, Joon-Young Lee, and In~So Kweon.
\newblock Cbam: Convolutional block attention module.
\newblock In {\em Proceedings of the European conference on computer vision
  (ECCV)}, pages 3--19, 2018.

\bibitem{xia2021adaptive}
Haifeng Xia, Handong Zhao, and Zhengming Ding.
\newblock Adaptive adversarial network for source-free domain adaptation.
\newblock In {\em Proceedings of the IEEE/CVF International Conference on
  Computer Vision}, pages 9010--9019, 2021.

\bibitem{yang2017learning}
Xiao Yang, Ersin Yumer, Paul Asente, Mike Kraley, Daniel Kifer, and C
  Lee~Giles.
\newblock Learning to extract semantic structure from documents using
  multimodal fully convolutional neural networks.
\newblock In {\em Proceedings of the IEEE Conference on Computer Vision and
  Pattern Recognition}, pages 5315--5324, 2017.

\bibitem{yu2019deep}
Zhou Yu, Jun Yu, Yuhao Cui, Dacheng Tao, and Qi Tian.
\newblock Deep modular co-attention networks for visual question answering.
\newblock In {\em Proceedings of the IEEE/CVF Conference on Computer Vision and
  Pattern Recognition}, pages 6281--6290, 2019.

\bibitem{zhang2019self}
Han Zhang, Ian Goodfellow, Dimitris Metaxas, and Augustus Odena.
\newblock Self-attention generative adversarial networks.
\newblock In {\em International conference on machine learning}, pages
  7354--7363. PMLR, 2019.

\bibitem{zhdanov2019diverse}
Fedor Zhdanov.
\newblock Diverse mini-batch active learning.
\newblock {\em arXiv preprint arXiv:1901.05954}, 2019.

\bibitem{zhong2019publaynet}
Xu Zhong, Jianbin Tang, and Antonio~Jimeno Yepes.
\newblock Publaynet: largest dataset ever for document layout analysis.
\newblock In {\em 2019 International Conference on Document Analysis and
  Recognition (ICDAR)}, pages 1015--1022. IEEE, 2019.

\bibitem{zhou2021discriminative}
Fan Zhou, Changjian Shui, Shichun Yang, Bincheng Huang, Boyu Wang, and Brahim
  Chaib-draa.
\newblock Discriminative active learning for domain adaptation.
\newblock {\em Knowledge-Based Systems}, 222:106986, 2021.

\end{thebibliography}
}

\end{document}